\documentclass[11pt,a4paper]{article}
\usepackage[hyperref]{emnlp2020}

\usepackage{times}
\usepackage{latexsym}
\usepackage{graphicx}
\usepackage{algpseudocode}
\usepackage{booktabs}
\usepackage{amsmath}
\usepackage{bbm}
\usepackage{cancel}
\usepackage[ruled,vlined]{algorithm2e}
\graphicspath{ {/} }

\usepackage{microtype}
\usepackage{xcolor,soul}
\usepackage{enumitem}

\usepackage{comment}
\usepackage[nameinlink]{cleveref}
\sethlcolor{pink}

\crefformat{section}{\S#2#1#3} 
\crefformat{subsection}{\S#2#1#3}
\crefformat{subsubsection}{\S#2#1#3}

\usepackage{amsmath,amsfonts,bm}

\newcommand{\mytilde}{{\raise.17ex\hbox{$\scriptstyle\mathtt{\sim}$}}}

\newcommand{\red}[1]{\textcolor{red}{#1}}
\newcommand{\blue}[1]{\textcolor{blue}{#1}}

\newcommand{\teal}[1]{\textcolor{teal}{#1}}
\newcommand{\brown}[1]{\textcolor{brown}{#1}}

\newcommand{\ie}{{\em i.e.,}}
\newcommand{\eg}{{\em e.g.,}}

\newcommand{\Ni}{({\em i})~}
\newcommand{\Nii}{({\em ii})~}

\newcommand{\Na}{({\em a})~}
\newcommand{\Nb}{({\em b})~}
\newcommand{\Nc}{({\em c})~}
\newcommand{\Nd}{({\em d})~}
\newcommand{\Ne}{({\em e})~}

\def\eqref#1{equation~\ref{#1}}

\def\1{\bm{1}}

\def\ry{{\textnormal{y}}}

\def\va{{\bm{a}}}

\def\vd{{\bm{d}}}

\def\vf{{\bm{f}}}

\def\vh{{\bm{h}}}

\def\vs{{\bm{s}}}
\def\vt{{\bm{t}}}

\def\vw{{\bm{w}}}
\def\vx{{\bm{x}}}

\def\mH{{\bm{H}}}

\def\mM{{\bm{M}}}

\def\mW{{\bm{W}}}
\def\mX{{\bm{X}}}

\DeclareMathAlphabet{\mathsfit}{\encodingdefault}{\sfdefault}{m}{sl}
\SetMathAlphabet{\mathsfit}{bold}{\encodingdefault}{\sfdefault}{bx}{n}

\def\gL{{\mathcal{L}}}

\def\gU{{\mathcal{U}}}

\newcommand{\softmax}{\mathrm{softmax}}
\newcommand{\sigmoid}{\mathrm{sigmoid}}

\DeclareMathOperator*{\argmax}{arg\,max}

\usepackage{microtype}

\aclfinalcopy 
\def\aclpaperid{1863} 

\title{Online Conversation Disentanglement with Pointer Networks}

\author{Tao Yu$^\P$ \and Shafiq Joty$^\P$$^\dagger$ \\
$^\P$Nanyang Technological University, Singapore \\
$^\dagger$Salesforce Research\\
{\tt \{tao003, srjoty\}@ntu.edu.sg}
\\}

\begin{document}
\maketitle
\begin{abstract}
Huge amounts of textual conversations occur online every day, where multiple conversations take place concurrently. Interleaved conversations lead to difficulties in not only following the ongoing discussions but also extracting relevant information from simultaneous messages. Conversation disentanglement aims to separate intermingled messages into detached conversations. However, existing disentanglement methods rely mostly on handcrafted features that are dataset specific, which hinders generalization and adaptability. In this work,
we propose an end-to-end online framework for conversation disentanglement that avoids time-consuming domain-specific feature engineering. We design a novel way to embed the whole utterance that comprises timestamp, speaker, and message text, and propose a custom attention mechanism that models disentanglement as a pointing problem while effectively capturing inter-utterance interactions in an end-to-end fashion. We also introduce a joint-learning objective to better capture contextual information.
Our experiments on the Ubuntu IRC dataset show that our method achieves state-of-the-art performance in both link and conversation prediction tasks. 

\end{abstract}



\section{Introduction}
With the fast growth of Internet and mobile devices, people now commonly communicate in the virtual world to discuss events, issues, tasks, and personal experiences. Among the various methods of communication, text-based conversational media, such as Internet Relay Chat (IRC), Facebook Messenger, Whatsapp, and Slack, has been and remains one of the most popular choices. Multiple ongoing conversations seem to occur naturally in such social and organizational interactions, especially when the conversation involves more than two participants \cite{elsner2010disentangling}. For example, consider the excerpt of a multi-party conversation in Figure \ref{fig:ubuntudata} taken from the Ubuntu IRC corpus \cite{lowe2015ubuntu}. Even in this small excerpt, there are 4 concurrent conversations (distinguished by different colors) among 4 participants.

\begin{figure}
\centering
\setlength{\tabcolsep}{3pt}
\scalebox{0.75}{\begin{tabular}{lll}
\toprule
Time & Sp & Message Text \\
\midrule
\brown{02:26} & \brown{system}& \brown{===\hl{zelot} joined the channel}\\
\teal{02:26} & \teal{zelot} & \teal{hi, where can i get some help in} \\ 
& & \teal{regards to issues with mount? }\\
\blue{02:26} & \blue{TuxThePenguin}&  \blue{After taking it out} \\
\blue{02:26} & \blue{hannasanarion}&  \blue{\hl{TuxThePenguin}, try booting with} \\
& &\blue{ monitors connected to motherboard }\\
\blue{02:26} & \blue{pnunn }& \blue{\hl{TuxThePonguin}, sounds like there is} \\
& & \blue{on board graphics as well, so try that}\\
& & \blue{without the card }\\
\blue{02:26} & \blue{pnunn} & \blue{Yeh, just one monitor though }\\

\blue{02:27} & \blue{hannasanarion} & \blue{\hl{pnunn}, right}\\
\blue{02:27} & \blue{TuxThePenguin} & \blue{Makes sense to me :) }\\
\blue{02:27} & \blue{pnunn} & \blue{process of elimination.} \\
\blue{02:27} & \blue{TuxThePenguin} & \blue{Along with Occam's Razor }\\
\teal{02:27} & \teal{Bashing-om} & \teal{\hl{zelot}: If you are on a supported release} \\
& & \teal{of 'buntu, this is a good place to ask.} \\
\blue{02:27} & \blue{TuxThePenguin} & \blue{Any solution is most likely the }\\
& & \blue{simplest one}\\
\red{02:28} & \red{wllrt} & \red{ I'm a emacs newb and looking to} \\
& & \red{prevent rsi.}\\ 
\bottomrule
\end{tabular}}
\caption{An excerpt of a conversation from the Ubuntu IRC corpus (best viewed in color). Same color reflects same conversation. Mentions of names are highlighted.} 
\label{fig:ubuntudata}
\end{figure}



Identifying or disentangling individual conversations is often considered as a prerequisite for downstream dialog tasks such as utterance ranking and generation \cite{lowe2017training,kim2019eighth}. It can also help building other applications such as search, summarization, and question answering over conversations, and support users by providing online help \cite{joty-etal-2019-discourse}.


However, often there are no explicit structures or metadata to separate out the individual conversations. 
Naive heuristics to disentanglement often lead to sub optimal results as \citet{acl19disentangle} found that only 10.8\% of the conversations in the widely used Ubuntu IRC dialog corpus were extracted correctly by the heuristics employed by \citet{lowe2015ubuntu,lowe2017training}.


Previous studies have therefore investigated traditional machine learning methods with statistical  and linguistic features for conversation disentanglement, \eg\  \cite{Shen:2006}, \cite{wang2009context,wang2011predicting,wang2011learning}, \cite{elsner2010disentangling,elsner-charniak-2011-disentangling}, to name a few. The task is generally solved by first finding links between utterances, and then grouping them into a set of distinct conversations. Recent work by \citet{jiang-etal-2018-identifying} and \citet{acl19disentangle} adopt deep learning approaches to learn abstract linguistic features and compute message pair similarity. However, these methods heavily rely on hand-engineered features that are often too specific to the particular datasets (or domains) on which the model is trained and evaluated. For example, many of the features used in \cite{acl19disentangle} are only applicable to the Ubuntu IRC dataset. This hinders the model's generalization and adaptability to other domains.

In this work, we propose a more general framework for conversation disentanglement while avoiding time-consuming and domain-specific feature engineering. In particular, we cast link prediction as a pointing problem, where the model learns to point to the parent of a given utterance (Figure \ref{fig:overview}). Each pointing operation is modeled as a multinomial distribution over the set of previous utterances. 
A neural encoder is used to encode each utterance text along with its speaker and timestamp. The pointing function implements a custom attention mechanism that models different interactions between two utterances. This results in an end-to-end neural framework that can be optimized with a simple cross entropy loss. During training, we jointly model the reply-to relationship and pairwise relationship (whether two utterances in the same conversation) under the same framework, so that more contextual and structural information can be learned by our model to further improve the disentangling performance.

Furthermore, the framework supports online decoding, which is naturally provided by the pointer network framework, disentangles a conversation as it unfolds, and can provide real-time help to participants in contributing to the right conversations.

We performed extensive experiments on the recently released Ubuntu IRC dataset \cite{acl19disentangle} and demonstrate that our approach outperforms previous methods for both link prediction and conversation prediction tasks.\footnote{\url{https://github.com/vode/onlinePtrNet_disentanglement}} Ablation studies reveal the importance of different components of the model and special handling of self-links. Our framework is generic and can be applied to chat conversations from other domains.

\section{Background}
In this section, we give a brief overview of previous work on conversation disentanglement and the generic pointer network model. 

\subsection{Conversation disentanglement}

Most existing approaches treat disentanglement as a two-stage problem. The first stage involves \emph{link prediction} that models ``reply-to'' relation
between two utterances. The second stage is a \emph{clustering} step, which utilizes the results from link prediction to construct the individual conversation threads.

For link prediction, earlier methods used discourse cues and content features within statistical classifiers. \citet{elsner-charniak-2008-talking,elsner2010disentangling} combine conversation cues like speaker, mention, and time with content features like the number of shared words to train a linear classifier. 

Recent methods use neural models to represent utterances with compositional features. \citet{mehri-carenini-2017-chat} pre-train an LSTM network to predict \emph{reply} probability of an utterance, which is then used in a link prediction classifier along with other handcrafted features. 
\citet{jiang-etal-2018-identifying}  model high and low-level linguistic information using a siamese hierarchical convolutional network that models similarity between pairs of utterances in the same conversation. The interactions between two utterances is captured by taking element-wise absolute difference of the encoded sentence features along with other handcrafted features.  
\citet{acl19disentangle} uses feed-forward networks with averaged pre-trained word embedding and many hand-engineered features. \citet{tan-etal-2019-context} used an utterance-level LSTM network, while \citet{zhu2019did} used a masked transformer to get a context-aware utterance representation considering utterances in the same conversation.


Finding a globally optimal clustering solution for conversation disentanglement has been shown to be NP-hard \cite{NIPS2004_2557}. Previous methods focus mostly on approximating the global optimal by either using greedy decoding \cite{wang2009context,elsner-charniak-2008-talking,elsner2010disentangling,elsner-charniak-2011-disentangling,jiang-etal-2018-identifying,aumayr2011reconstruction} or training multiple link classifiers to do voting \cite{acl19disentangle}. \citet{mehri-carenini-2017-chat} trained additional classifiers to decide whether an utterance belongs to a conversation or not. \citet{Weishi-et-al-emnlp-20} use a multi-task topic tracking framework for conversation disentanglement, topic prediction and next utterance ranking. 

Our work is fundamentally different from previous studies in that we treat link prediction as a pointing problem modeled by a multinomial distribution over the previous utterances (as opposed to pairwise binary classification). This formulation allows us to model the global conversation flow. Our method does not rely on any handcrafted features. Each utterance in our method is represented by the utterance text, its speaker and timestamp, which are generic to any conversation. The interactions between the utterances are effectively modeled within the pointer module. Moreover, our framework can work in an end-to-end online setup.




\subsection{Pointer Networks}

Pointer networks \cite{vinyals2015pointer} are a class of encoder-decoder models that can tackle problems where the output vocabulary depends on the input sequence. They use attentions as pointers to the input elements. An encoder network first transforms the input sequence $\mX = (\vx_1, \ldots, \vx_n)$ into a sequence of hidden states $\mH = (\vh_1, \ldots, \vh_m)$. At each time step $t$, the decoder takes the input from the previous step, generates a decoder state $\vd_t$, and uses it to attend over the input elements. The attention gives a $\softmax$ (multinomial) distribution over the input elements as follows.
\begin{equation}
s_{t,i} = \sigma(\vd_t, \vh_i); \hspace{1em}
\va_t = \softmax(\vs_t)
\label{eq:attn}
\end{equation}
where $\sigma(.,.)$ is a scoring function for attention, which can be a neural network or simply a dot product operation. The model uses $\va_t$ to infer the output: $\hat{\ry}_t = \argmax (\va_t)$.


Similar to the standard pointer network, each pointing mechanism in our approach is modeled as a multinomial distribution over the indices of the input sequence. However, unlike the original pointer network where a decoder state points to an encoder state, in our approach, the current encoder state points to the previous states.

Pointer networks have recently yielded state-of-the-art results in constituency parsing \cite{Nguyen-et-al-acl-20}, dependency parsing \cite{ma-etal-2018-stack}, anaphora resolution \cite{lee2017anaphora}, and discourse segmentation and parsing \cite{lin-etal-2019-unified}. To the best of our knowledge, this is the very first work that utilizes a pointer network for conversation disentanglement. It is also a natural fit for online conversation disentanglement.



\section{Our Disentanglement Model}

Given a sequence of streaming utterances $\gU = \{U_1,U_2,\ldots, U_i, \ldots \}$, our task in link prediction (\cref{subsec:point}) is to find the parent utterances $U_{p_i} \subset U_{\le i}$  that the current utterance $U_i$ replies to. Here, $U_{\le i}$ refers to all the  previous utterances until $i$, that is, $U_{\le i} = (U_0, U_1\ldots, U_{i})$. An utterance can reply to itself (\ie\ $i = p_i$), for example, the initial message in a conversation or a system message. Besides, one utterance may have multiple parents, and one parent can be replied to by multiple (children) messages. For example, in \Cref{fig:ubuntudata}, both \texttt{pnunn} and \texttt{hannasanarion} reply to \texttt{TuxThePonguin}'s message. The case of one message replying to multiple parents is very rare in our corpus (see \Cref{table:overall_statistic}).

After link prediction, we employ a decoding algorithm (\cref{subsec:decoding}) to construct the individual threads. 




\subsection{Link Prediction by Pointing} \label{subsec:point}

 

\begin{figure*}[t!]
    \centering
    \includegraphics[scale= .2]{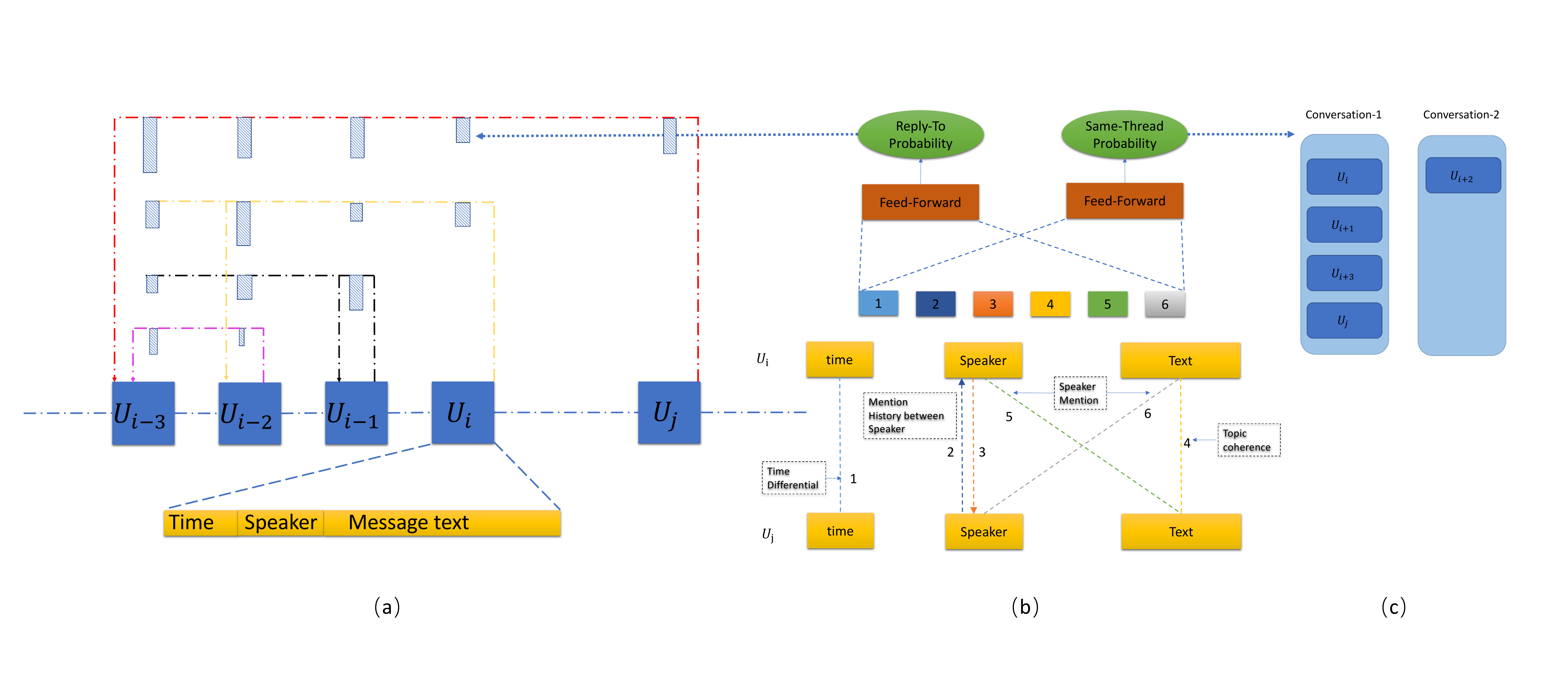}
    \caption{(a) Overview of our Pointer Network joint learning framework for online conversation disentanglement. Each utterance $U_i$ consists of three parts: time ($t_i$), speaker ($s_i$) and message text ($m_i$), which are encoded by the utterance encoder. (b) The Pointer module designs an effective attention mechanism that captures inter-utterance interactions through several features and models link prediction as a multinomial distribution over the previous utterances. (c) The pairwise classification model aims to capture higher-order contextual information. The whole model is trained end-to-end and can decode/disentangle the conversation in an online fashion.} 
    \label{fig:overview}
\end{figure*}

We propose a joint learning  framework for conversation disentanglement based on pointing operations where \Cref{fig:overview} shows the network architecture. It has three main components: \Na an utterance encoder and \Nb a pointer module \Nc a pairwise classification model. The job of the utterance encoder is to encode each utterance $U_i$ as it comes, while the the pointer module implements a custom pointing mechanism to find the ancestor message $U_p \in U_{\le i}$ that $U_i$ replies to. The pairwise classification model is to determine whether two utterances $U_i$ and $U_j$ are in the same conversation or not.

\subsubsection{Utterance Encoder}

As shown in \Cref{fig:ubuntudata}, each utterance $U_i$ has three components $<$$t_i,s_i,m_i$$>$: the {timestamp} $t_i$ when the utterance was posted, the {speaker} $s_i$ who posted it, and the {message content} $m_i$. We encode these three components separately.

\begin{description}[leftmargin=0pt]

\item [Encoding timestamp.] The timestamp \texttt{hour}:\texttt{min} is directly encoded as a two-dimensional vector $[hour, min]$. For example, timestamp \texttt{02:26} is encoded as $[02,26]$. 

\item [Encoding speaker. ] 
In a multi-party conversation, participants mention each other's names to make disentanglement easier, compensating for the lack of visual cues normally present in a 
face-to-face conversation \cite{ONeill:2003,elsner2010disentangling}. Our goal is to capture the mention relation between utterances in the way we encode the speaker information. {For this, each speaker is placed in the same vocabulary as the words, and encoded with a unique identifier (a discrete value).}

\item [Encoding message text.] We utilize a Bidirectional LSTM  or Bi-LSTM  \cite{hochreiter1997long} to encode the raw text message $m_i$ into deep contextual representations of the words. We concatenate the hidden states from both forward and backward  LSTM cells. Formally, for a message containing $n$ words $m_i = (w_0,w_1,\ldots,w_n)$, the Bi-LSTM gives $\mH = (\overset{\rightarrow}{\vh_0}\oplus \overset{\leftarrow}{\vh_0},\overset{\rightarrow}{\vh_1}\oplus \overset{\leftarrow}{\vh_1},\ldots,\overset{\rightarrow}{\vh_n}\oplus \overset{\leftarrow}{\vh_n})$, where $\oplus$ denotes vector concatenation.

\end{description}

\subsubsection{Pointer Module}
\label{subsubsec:pointer}

Given the encoded representation of the current utterance $U_i$, our pointer module computes a probability distribution over the previous utterances $U_{\le i}$ which represents the probability that $U_i$ replies to an utterance $U_{j} \in U_{\le i}$. The module implements an association function between $U_i$ and $U_j$ by incorporating different kinds of interactions. \Cref{fig:overview}(b) gives a schematic diagram of the module, which  has five different components:



\begin{description}[leftmargin=0pt]

\item [\Na Time difference:] The time difference  between $U_i$ and $U_j$ is computed as: $\vt_{i,j} = [\vt_{i} - \vt_{j}]$.

\item[\Nb Mention:]  To determine whether $U_i$ mentions $s_j$ (speaker of $U_j$) in its message $m_i$, we compute:
\begin{equation}
        {mention}_{i,j} = \sum_{w_k \in m_i} \mathbbm{1} (s_{j} = \text{Index}(w_{k})) 
\end{equation}
\noindent where $\mathbbm{1}$ is the Indicator function that returns $1$ if the index of $w_k$ in $m_i$ matches the speaker id $s_j$.

\item[\Nc Mention history:] The pointer module also keeps track of mention histories between two speakers. It not only computes whether $s_i$ mentions $s_j$ in $m_i$, but it also keeps track of whether $s_i$, $s_j$ mentioned each other in their previous messages and how often. It maintains an external memory $\mM$ (a matrix) to record this. At each step, we compute both ${mention}_{i,j}$ and ${mention}_{j,i}$, and update the memory $\mM$ to be used in the next step.   
\begin{eqnarray}
        \mM_{i,j} = \mM_{i,j} + \text{mention}_{i,j} \\
        \mM_{j,i} = \mM_{j,i} + \text{mention}_{j,i} 
\end{eqnarray}
The memory $\mM$ grows incrementally as the model sees new speakers during training and inference.

\item[\Nd Topic coherence: ] To model textual similarity between $m_i$ and $m_j$, we use a similar method as \citet{chen2016enhanced}. Let $\mH_i = (\vh_{i,0}, \ldots, \vh_{i,p})$ and $\mH_j = (\vh_{j,0}, \ldots, \vh_{j,q})$ be the Bi-LSTM representations for $m_i$ and $m_j$, respectively from the utterance encoder layer. We compute soft alignment between $m_i$ and $m_i$ as follows.
\begin{eqnarray}
        \mH_i^{'} = \softmax(\mH_i \mH_j^T) \mH_j  \label{eq:qhi}  \\
        \mH_j^{'} = \softmax(\mH_j \mH_i^T) \mH_i 
\label{eq:qhj}
\end{eqnarray}
In Eq. \ref{eq:qhi}, we use the vectors in $\mH_i$ as the query vectors to compute attentions over the key/value vectors in $\mH_j$, and compute a set of attended vectors $\mH_i^{'} = (\vh_{i,0}^{'}, \ldots, \vh_{i,p}^{'})$, one for each $\vh_i \in \mH_i$. Eq. \ref{eq:qhj} does the same thing but uses $\mH_j$ as the query vectors and $\mH_i$ as the key/value vectors to compute the attended vectors, $\mH_j^{'} = (\vh_{j,0}^{'}, \ldots, \vh_{j,q}^{'})$.



Then we enhance the interactions by applying difference and  element-wise product between the original representation $\mH$ and the attended representations $\mH^{'}$ as follows.
\begin{eqnarray}
    \vh_i^{f} &=& [\vh_i; \vh_i^{'}; \vh_i - \vh_i^{'}; \vh_i \cdot \vh_i^{'}] \\
    \vh_j^{f} &=& [\vh_j; \vh_j^{'}; \vh_j - \vh_j^{'}; \vh_j \cdot \vh_j^{'}] 
\end{eqnarray}

The final representation $\vh_{i,j}$ is computed as 
\begin{equation}
    \vh_{i,j} = \vh_i^{f} \oplus \vh_j^{f}
    \label{eq:esim_final}
\end{equation}
where $\oplus$ denotes concatenation. 
\item[\Ne Pointing: ] After computing the above four types of interactions between each pair of utterances, we concatenate them and feed them into a feed-forward network to compute the pointing distribution over all the previous utterances. 
\begin{eqnarray}
    \vf_{i,j}  =  &\vt_{i,j} \oplus {mention}_{i,j}  \oplus {mention}_{j,i} \nonumber \\ 
    & \oplus \mM_{i,j} \oplus \mM_{j,i} \oplus \vh_{i,j} 
    \label{eq:feature}
\end{eqnarray}
\begin{eqnarray}
    \text{score}(U_{i}, U_{j})  = \tanh (\vw^T  \vf_{i,j}) \nonumber \label{eq:score} \\  
  p(U_i, U_j) =   \frac{\exp(\text{score}(U_i,U_j))}{\sum_{s=0}^{i}\exp(\text{score}(U_i,U_s)) }  \label{eq:point} \\ 
  \hspace{3em} \text{for } j \in (0,...,i) \nonumber 
\end{eqnarray}
\end{description}
where $\vw$ is a shared linear layer parameter. We use cross entropy (CE) loss for the pointer module. 
\begin{equation}
\gL_{\text{link}} (\theta) = -\sum_{j=0}^{i} y_{i,j} \log p(U_i, U_j)
\label{eq:loss}
\end{equation}
\noindent where  $y_{i,j} = 1$ if $U_i$ replies to $U_j$, otherwise $0$, and $\theta$ are the model parameters.

\subsubsection{Pairwise Classification Model}

In the above pointer module, we only consider first-order interaction between two utterances in an online fashion (\ie\ looking at only previous utterances). However, higher-order information derived from the entire conversation may provide more contextual information for the model to learn useful disentanglement features. We propose a joint learning framework to consider both first- and higher-order information simultaneously in a unified framework; see Figure \ref{fig:overview}(b)-(c).

For the higher-order information, we train a binary pairwise classifier that decides whether two utterances should be in the same conversation. For any two arbitrary utterances we use the same feature function from Eq. \ref{eq:feature} (\ie\ the parameters are shared with the pointer module) and feed them into a binary logistic classifier. The probability of two utterances belongs to the same conversation is:
\begin{eqnarray}
    p_{\text{pair}}(U_i,U_j)  = \sigmoid (\vw^T \vf_{i,j}) \label{eq:cluster_prob}
\end{eqnarray}
\noindent where $\vw$ is the classifier parameter. We use a binary cross entropy loss for this model. 
\begin{equation}
\small
\gL_{\text{pair}}(\theta) = - y_{i,j} \log \hat{y}_{i,j} - (1 - y_{i,j}) \log (1-\hat{y}_{i,j})
\label{eq:loss}
\end{equation}
\noindent where $\hat{y}_{i,j} = p_{\text{pair}}(U_i, U_j)$, $y_{i,j} = 1$ if $U_i$ and  $U_j$ are in the same conversation, otherwise $0$, and $\theta$ are the model parameters. Since the pairwise classifier is trained on all possible pairs of utterances in a conversation, it models higher-order information about the conversation clusters.

\subsection{Training}

The final training loss of our model is: 
\begin{equation}
\gL (\theta) = \gL_{\text{link}} (\theta) + \lambda\gL_{\text{pair}} (\theta) 
\label{eq:loss}
\end{equation}
\noindent where  $\lambda$ is the hyper-parameter for tuning the importance of the pairwise classification loss. We use Glove 128-dimensional word embedding \cite{pennington-socher-manning:2014:EMNLP2014},  pre-trained by  \citet{lowe2015ubuntu} on the \#Ubuntu corpus. The hidden layers in the Bi-LSTM are of 256 dimensions. We optimize our model using Adam \cite{kingma2014adam} optimizer with a learning rate of $1 \times 10^{-5}$. For regularization, we set the dropout at $0.2$ and $L_2$ penalize weight with $1 \times 10^{-7}$.

\subsection{Decoding} \label{subsec:decoding}


Our framework naturally allows us to disentangle the threads in an online fashion. As a new utterance $U_i$ arrives, the utterance encoder encodes it into a vector. The pointer module then computes a multinomial distribution over the previous utterances $U_{\le i}$ (Eq. \ref{eq:point}) by modelling pair-wise interactions, and then finds the parent message $U_{p_i}$ as follows.
\begin{equation}
    p_i = \argmax_{j=\{0,\ldots, i\}} ~~p(U_i, U_j)
\end{equation}

To the best of our knowledge, this is the first work using an end-to-end framework for online conversation disentanglement.  In our analysis of several conversations, we found that the self-links (an utterance pointing to itself) play a crucial role in clustering performance. Mistakes in correctly identifying a self-link will result in two misclusterings. To address this, we did some simple adjustment to our decoding method. We raise the threshold for self-link prediction to make a more conservative prediction of self-links.  
In particular, during decoding we first find the parent with the highest probability, but if it turns out to be a self-link, we see if the probability passes the preset threshold, otherwise the utterance with the second highest probability will be predicted as the parent. The tuning of the threshold parameter for self-link is done on the development set.

\section{Experiment}

In this section, we present our experiments --- the dataset used, the evaluation metrics, experimental setup, and the results with analysis. 

\subsection{Dataset}

The dataset used for training and evaluation is from \cite{kim2019eighth}, which is the largest dataset available for conversation disentanglement \cite{acl19disentangle}. It consists of multi-party conversations extracted from the \#Ubuntu IRC channel. 
A typical conversation starts with a question that was asked by one participant, and then other participants respond with either an answer or follow-up questions. This leads to a back-and-forth conversation between multiple participants. An example of the Ubuntu-IRC data is shown in  \Cref{fig:ubuntudata}. We follow the same train, dev, test split as the Dialog System Technology Challenges 8 (DSTC8)  \cite{kim2019eighth}.\footnote{The ground truth for test set can be found at \url{http://jkk.name/irc-disentanglement/}}  Table \ref{table:overall_statistic} reports the dataset statistics.

\begin{table}[t!]
\scalebox{0.80}
{\begin{tabular}{l|ccc}
\toprule
& Train & Dev & Test \\
\midrule
 Total \# links  & 52641 & 2145 & 4265 \\
 Total \# conversations  & 6201 & 526 & 370 \\
 Avg link distance & 8& 8 & 7 \\
 Median link distances & 3 & 3 & 3 \\
 Avg parents per utterance & 1.03& 1.05 & 1.04 \\
 Avg \# of utterances per conv. & 9& 5 & 12\\
 Median \# of utterances per conv. &5 & 4 &6\\
 \bottomrule
\end{tabular}}
\caption{Statistics of train, dev and test datasets.}
\label{table:overall_statistic}
\end{table}

\subsection{Metrics}

We consider two kinds of metrics to evaluate our disentanglement model: link level and conversation level. For link-level, we use  precision, recall and F-1 scores. For cluster level evaluation, we use the same clustering metrics from DSTC8 and \cite{acl19disentangle}. This includes: 

\begin{description}[leftmargin=0pt]
    \item [\Na Variation of Information (VI). ] This is a measure of information gain or loss when going from one clustering to another \cite{meilua2007comparing}. It is the sum of conditional entropies $H(Y |X) + H(X|Y )$, where $X$ and $Y$ are clusterings of the same set of items. We used the bound for $n$ items that $VI(X; Y ) \le \log (n) $, and present $1 - VI$, so that the larger the value the better.
    
    \item [\Nb Ajusted Random Index (ARI).] A measure (also referred to as \textbf{1V1})  \cite{hubert1985comparing}  between two clusterings by considering all links of samples and counting links that are assigned in the same or different clusters in the predicted and true clusterings. ARI is defined as:
\begin{equation}
\small
    \frac{\sum_{ij} \binom{n_{ij}}{2} - [\sum_i \binom{a_i}{2} \sum_j \binom{b_j}{2}] / \binom{n}{2}}{\frac{1}{2} [\sum_i \binom{a_i}{2} + \sum_j \binom{b_j}{2}] - [\sum_i \binom{a_i}{2} \sum_j \binom{b_j}{2}] / \binom{n}{2}}
    \nonumber
\end{equation}
\normalsize

\noindent where $n_{ij}$ are number of overlapping links between predicted cluster $i$ and ground truth cluster $j$, whereas  $a_i$ and $b_j$ indicate row and column level summation over $n_{ij}$.

\item [\Nc Exact Match F-1.] Calculated using the number of perfectly matching conversations, excluding conversations with only one message.

\end{description}


\begin{table*}[t!]
\centering
\scalebox{0.86}{{\begin{tabular}{l|ccccc|ccc|ccc}
\toprule
Model& \multicolumn{5}{c}{\textbf{Cluster Prediction}} & \multicolumn{3}{|c|}{\textbf{Link Prediction}} &
\multicolumn{3}{|c}{\textbf{Self-Link Prediction}} \\
 \midrule
  & VI & 1V1 & P & R & F1 & P & R & F & P & R & F1 \\
 \midrule
 FF (T) & 66.7 & 10.0 & 0.4 & 0.5 & 0.7 & 19.7 & 19.0& 19.4 & 23.0& 98.0& 60.5\\ 
 Ptr-Net (T)  & 69.3 & 22.5 & 2.6 & 4.2 & 3.2& 25.0& 24.0 & 24.5 & 60.2& 64.3 & 62.3 \\ 
 \midrule
 FF ($-$T) & 90.2 & 62.1 & 26.8 & 32.3 & 29.3 & 71.3& 68.7& 70.0 & 80.1 & 91.0& 85.5 \\ 
 Ptr-Net (-T) & 89.5 & 61.0 & 27.3 & 29.5 & 28.4 & 71.0& 67.2 & 69.2 & 78.2 & 90.0 & 84.1 \\ 
 \midrule
 SHCNN & 87.1& 62.3 &20.9 & 31.0 & 25.1 & 71.7 & 69.1& 70.4 & 79.5 & 80.0 & 80.3 \\
 \midrule
 FF& 92.2 & 69.6 & 38.7& 41.6 & 40.1& 74.5& 71.8& 73.1 & 86.2& 92.5& 89.4\\ 
 + Self-Link& 92.0 & 70.4 & 41.9 & 40.1 & 41.0  & 74.2& 71.5& 72.8 & 92.1& 90.0& 91.0 \\ 
 \midrule
 Ptr-Net  & 92.3 & 70.2 & 33.0 & 38.9 & 36.0 & 74.7& 72.7 & 73.7 & 81.8& 92.1 & 87.0 \\ 
 + Joint train  & 93.1 & 71.3 & 37.2 & 42.5 & 39.7 & 74.0& 71.3 & 72.7 & 79.5& \textbf{93.6} & 86.6 \\ 
+ Self-link  & 93.0 & 74.3 & 42.2 & 40.9 & 41.5& \textbf{74.8}& \textbf{72.7} & \textbf{73.7} & 92.2& 89.4 & 91.3\\ 
 + Joint-train \& Self-link & \textbf{94.2} & \textbf{80.1} & \textbf{44.9} & \textbf{44.2} & \textbf{44.5}& 74.5& 71.7 & 73.1 & \textbf{92.8}& 90.2 & \textbf{91.5}\\ 
 \bottomrule
\end{tabular}}}
\caption{Experimental results on the Ubuntu test set. ``T" suffix means the model uses only utterance text. ``$-$T" indicates the model excludes utterance text. ``Joint Train" indicates the model is trained with the joint learning objective (Eq. \ref{eq:loss}), ``Self Link" indicate the model is decoded with self-link threshold re-adjustment.}
\label{table:experiment_result}
\end{table*}

\subsection{Models Compared}

\paragraph{Feed Forward Model.}  We use the feed-forward model from \cite{acl19disentangle} as the baseline model, which outperforms previously proposed disentanglement models. For the DSTC8 challenge, the author (one of the task organizers) provided a trained model\footnote{\url{https://github.com/dstc8-track2/NOESIS-II/tree/master/subtask4}}, which has two feed-forward layers. The input is 77 hand-engineered features combined with 128 dimension word average embeddings from pre-trained Glove. We will denote this model as FF model below.

\paragraph{Pointer Network.} This is our model. For computational simplicity, we did not compute the attention over all the previous utterances, rather we set a fixed window size of $50$. This means for the current utterance, we will calculate the attention with itself and $50$ previous utterances during training and decoding. In our training data, about 97\%  of the utterances' parents are located in this window. In the \#Ubuntu Data,  according to the statistic of \Cref{table:overall_statistic}, one utterance only have 1.03 parent utterances on average. So, given an utterance we only predict its most likely parent.

\subsection{Results}

We present our main results in \Cref{table:experiment_result}. For analysis purposes, in the table we also show the results for two variants of the models: \Ni when the models consider only the utterance texts, as denoted by (T) suffix; \Nii when the models exclude the utterance text, as denoted by ($-$T) suffix. In addition, we present  how the models perform specifically on  self-links predictions, as correctly identifying self-links turns out to be quite crucial for identifying the conversations, as we will explain later. {We also report the performance of  the Siamese hierarchical convolutional neural network (SHCNN) from \citep{jiang-etal-2018-identifying} on \#Ubuntu dataset. However, SHCNN mainly focuses on modeling message content representations and only incorporates four context features: speaker identicality, absolute time difference, and the number of duplicate words. So the performance is not as good as the feed-forward model with many hand-engineered features.}


\paragraph{Link Prediction.} We can  see that our Pointer Network has better \textbf{link prediction} accuracy compared to the baseline, when it uses the message texts. The reason that the baseline performs slightly better in the absence of message texts is because it uses several meta features from the whole thread that capture more structural information. On the other hand, our model has access to only time and speaker information in the absence of message texts.  Thus, we can say that our model can capture textual similarity or topical coherence better compared to the baseline.

\paragraph{Cluster (or Conversation) Prediction.}
Now if we compare the performance at the \textbf{cluster-level}, we see that our model performs much better when it uses only textual information compared to the baseline. In the absence of textual information, it performs on par with the baseline. However, when we compare the full model, we notice that its results are lower in some cluster-level measures (see `Ptr-Net' results in the last block), which we did not expect given that it has higher accuracy on link prediction. Therefore, we performed a case study, and the study reveals that one particular kind of links, which we call ``self-link'', are very crucial to the cluster-level results.

\begin{table}[t!]
\centering
{\begin{tabular}{lc}
\toprule
 Link Type & Percentage \\
 \midrule
 System Messages & 41\% \\
 Start of Conversation & 36\% \\
 Isolated Messages & 33\% \\
 \bottomrule
\end{tabular}}
\caption{Self-link statistics on Ubuntu Dataset}
\label{table:self_link_statistic}
\end{table}

\paragraph{Self-links. } \citet{acl19disentangle} mention that most of the self-links are system messages like  \texttt{"===zelot just join the channel"}. However, according to our statistics (shown in \Cref{table:self_link_statistic}), only 41\% self-links are system messages, and we have identified two other types of utterances that reply to themselves:
\begin{itemize}[leftmargin=*]  
    \item \textit{Start of a conversation:} These messages do not reply to any previous message but will be replied afterward. 
    \item \textit{Isolated Messages:} These are non-system messages, but reply to no previous message and never been replied afterwards.
\end{itemize}

\begin{table}[t!]
\centering
\setlength{\tabcolsep}{3pt}
\scalebox{0.80}
{\begin{tabular}{l|ccccc|ccc}
\toprule
Model& \multicolumn{5}{|c|}{\textbf{Cluster} } & \multicolumn{3}{|c}{\textbf{Link}}\\
\midrule
& VI & 1V1 & P & R & F1 & P & R & F1 \\
 \midrule
FF& 92.2 & 79.3 & 38.7& 41.6 & 40.1& 74.5& 71.8& 73.1 \\
FF +G & 93.6 & 76.4 & 53.4 & 53.5 & 53.5 & 77.0& 74.2 & 75.6 \\
\midrule
Ptr-Net  & 92.3 & 70.2 & 33.0 & 38.9 & 36.0 & 75.4 & 72.7 & 74.1\\
Ptr-Net +G  & 94.5 & 79.3 & 55.1 & 54.7 & 54.9 & 78.4& 75.6 & 77.0 \\ 
 \bottomrule
\end{tabular}}
\caption{``+G" indicates replacing self-link prediction with ground truth labels.}
\label{table:self_link_groud_experiment}
\end{table}

\paragraph{Handling Self-links. } 
To see how much our models get affected by inaccurate self-link predictions, we performed an experiment using ground truth self-links, where we replace those predictions with ground truth self-link. From  \Cref{table:self_link_groud_experiment}, we see that although it shows only 3\% improvement on overall link-prediction for our model, but for cluster prediction, it increases by 13\% F1 for FF model and 19\% for our model. The reason is if a self-link is predicted wrong and the utterance links to another conversation,  it may destroy two true positive clusters. So the performance on self-link prediction could be a bottleneck for the clustering task.

\begin{table}[t!]
\centering
\setlength{\tabcolsep}{3pt}
{\begin{tabular}{llll}
\toprule
 Link Type & P & R & F1 \\
 \midrule
 System Messages & 99.0 & 100& 99.5 \\
 Start of Topic & 75.0 & 80.0 &77.5 \\
 Isolated Messages & 66.7 & 80.0 & 73.3 \\
 \bottomrule
\end{tabular}}
\caption{Self-link prediction results for the baseline model \cite{acl19disentangle}.}
\label{table:self_link_result}
\end{table}

\Cref{table:self_link_result} 
gives the detailed self-link prediction results for the FF model. This shows that the model can  predict the system messages with almost 100\% accuracy, but for other kinds of self-links, the performance is not that high. Experimental results in \Cref{table:experiment_result} show that our proposed Pointer Network has worse results on self-links compared to the FF model (compare ``FF" and `` Ptr-Net'' in the forth and fifth block), which explains why our model does not perform well on clustering metrics.

When we do the simple adjustment to our decoding method by raising the threshold for self-link prediction, we see significant improvements in clustering. Note that this adjustment also improves the performance of the baseline (FF+Self-Link).

\paragraph{Joint Learning. }
The results in \Cref{table:experiment_result} show that joint learning  further improves the clustering result. Combined with joint training, our online decoding algorithm achieves state-of-the-art results.

\paragraph{Ablation Studies.}

To show the importance of encoding textual and non-texture features like speaker and time, we trained the models with only textual features and without textual features; see the (T) ($-$T) variants in \Cref{table:experiment_result}. 

Intuitively, it is hard for the models to come up with a good prediction with only textual features, since the utterances in the same conversation usually talk about similar topics. This makes it difficult for the models to identify the right parent. Therefore, the  results in  \Cref{table:experiment_result} show a huge drop in performance when no speaker and time information are used and only textual features are used (see the first block of results with (T) suffix). This indicates that time and speaker mention information play a crucial role in disentanglement.

Similarly, the performance also goes down for the models when they do not consider textual information; see ($-$T) variants in \Cref{table:experiment_result}. Although compared to the text only, the drop is less.

\section{Conclusion}
In this paper, we have proposed a novel online framework for disentangling multi-party conversations. 
In contrast to previous work, our method reduces the effort of complicated feature engineering by proposing an utterance encoder and a pointer module that models inter-utterance interactions. Moreover, we propose a joint-training framework that enables the pointer network to learn more contextual information. Link prediction in our framework is modeled as a pointing function with a multinomial distribution over previous utterances. We also show that our framework supports online decoding. Extensive experiments have been conducted on the \#Ubuntu dataset, which show that our method achieves state-of-the-art performance on both link and conversation prediction tasks without using any handcrafted features.

There are some possible future directions from our work. We have shown in our experiments that self-link predictions have a significant impact on clustering results. This reminds us that neither our and most of the existing methods took good advantage of graph information in disentangling conversations. This can be done in two ways, encoding, and decoding. From the encoding side, it would be ideal to encode an utterance within its context. One challenge for this problem is that conversations are tangled, so sequential encoding methods like the one of \cite{sordoni2015hierarchical} would not be appropriate. From the decoding side, a promising direction would be to make global inference in a more efficient way.

\bibliographystyle{acl_natbib}
\bibliography{anthology,emnlp2020,acl2020}

\end{document}